# Reason induced visual attention for explainable autonomous driving

**Sikai Chen***
Visiting Assistant Professor, Center for Connected and Automated Transportation (CCAT), and Lyles School of Civil Engineering, Purdue University, West Lafayette, IN, 47907.
Email: chen1670@purdue.edu; and
Visiting Research Fellow, Robotics Institute, School of Computer Science, Carnegie Mellon University, Pittsburgh, PA, 15213.
Email: sikaichen@cmu.edu
ORCID #: 0000-0002-5931-5619
(Corresponding author)

**Jiqian Dong**
Graduate Research Assistant, Center for Connected and Automated Transportation (CCAT), and Lyles School of Civil Engineering, Purdue University, West Lafayette, IN, 47907.
Email: dong282@purdue.edu
ORCID #: 0000-0002-2924-5728

**Runjia Du**
Graduate Research Assistant, Center for Connected and Automated Transportation (CCAT), and Lyles School of Civil Engineering, Purdue University, West Lafayette, IN, 47907.
Email: du187@purdue.edu
ORCID #: 0000-0001-8403-4715

**Yujie Li**
Graduate Research Assistant, Center for Connected and Automated Transportation (CCAT), and Lyles School of Civil Engineering, Purdue University, West Lafayette, IN, 47907.
Email: li2804@purdue.edu
ORCID #: 0000-0002-0656-4603

**Samuel Labi**
Professor, Center for Connected and Automated Transportation (CCAT), and Lyles School of Civil Engineering, Purdue University, West Lafayette, IN, 47907.
Email: labi@purdue.edu
ORCID #: 0000-0001-9830-2071

**ABSTRACT**
Deep learning (DL) based computer vision (CV) models are generally considered as black boxes due to poor interpretability. This limitation impedes efficient diagnoses or predictions of system failure, thereby precluding the widespread deployment of DLCV models in safety-critical tasks such as autonomous driving. This study is motivated by the need to enhance the interpretability of DL model in autonomous driving and therefore proposes an explainable DL-based framework that generates textual descriptions of the driving environment and makes appropriate decisions based on the generated descriptions. The proposed framework imitates the learning process of human drivers by jointly modeling the visual input (images) and natural language, while using the language to induce the visual attention in the image. The results indicate strong explainability of autonomous driving decisions obtained by focusing on relevant features from visual inputs. Furthermore, the output attention maps enhance the interpretability of the model not only by providing meaningful explanation to the model behavior but also by identifying the weakness of and potential improvement directions for the model.

## INTRODUCTION

Results from recent socioeconomic surveys indicate that the most profound public concerns regarding autonomous vehicle (AV) adoption are safety, reliability, and trust in automation (*1–3*). In this regard, the success of an AV systems hinges on the coordination or cooperation of multiple modules of the system which include perception, motion planning, and control (*4*). However, the perception module is often considered the Achilles heel of the overall system as it has been the underlying cause of several AV-related fatal accidents in the recent years (*5*)(*6*). Because perception is the initial block encountered with the AV operations process, any error at this stage not only cascades but also may be amplified in subsequent stages of the autonomous driving process. Further, most of the current state-of-the-art (SOTA) algorithms in AV perception rely on deep learning (DL)-based computer vision (CV) models (*7–10*), benefiting from the merits of DL but also inheriting the intrinsic drawbacks. One of the biggest problems for DLCV models is that the black-box computation is notoriously inexplicable. Limited explainability has the effect that model failures are often undiagnosable and unpredictable, and this situation could ultimately exacerbate user distrust of automation. For this reason, the adoption of DL models in real-world AV systems still remains immature despite the acclaimed success of DL-based CV models in other domains. Therefore, this study is based on the notion that such interpretability problems could be mitigated by developing explainable DL systems. Such systems are capable of yielding outputs and intermediate results that can be understood by humans and therefore can be used by developers for "system debugging." One way to greatly enhance the explainability of model behavior is through the application of **attention mechanisms** and visualizing the attention maps (*11–13*).

### A. Background in Attention Mechanism

In DL, attention is a mechanism by which a neural network can automatically distinguish the relative importance of the features and weigh the features for downstream computations. Originally proposed by Bahdanau et al. (*14*), attention mechanism has been widely applied in the Natural Language Processing (NLP) domain for sequence-to-sequence (or "seq2seq") modeling such as speech recognition (*15*), machine translation (*16*), and other complex language tasks (*17, 18*). For seq2seq modeling, a key challenge is that there often exists complicated dependency between a specific part of the input sequence and a completely different part in the output sequence. For example, in machine translation, the corresponding words of a sentence with the same meaning in different languages may appear at completely different locations. Therefore, there is a need for the model to flexibly learn the dependency mechanism that can automatically "attend" to a specific position of the input sequence when generating the output sequence. As pointed out by Bahdanau et al. (*14*), the attention mechanism not only boosts the model performance by "enlarging" the useful features while "filtering out" the noise or redundant information, but also enhances the interpretability of the model by providing visualizable "attention maps". These attention maps can capture the semantic correspondence between input and output, thereby enabling more profound model diagnosis for a wider variety of DL models.

In CV tasks, there also exists a similar input-output dependency. However, instead of the dependency occurring along the temporal axis as is the case in natural language (that is, determining which word is more important), it occurs along the spatial domain (that is, determining which region is more important). It is vital for CV models to learn the relative importance of certain spatial regions in its input through visual attention as images generally contain significant redundant information. For example, with its broad receptive field of vision, an AV's onboard camera can capture irrelevant and frequently recurring information such as advertisement boards, far-off buildings, the sky, and so on. Having to process large amounts of redundant and irrelevant information not only constitutes a waste of computational resources but also reduces the model performance due to the added noise. On the other hand, human vision, as a natural filter, can quickly and effortlessly locate objects of visual interest and relevance while filtering out irrelevant details in the image (*19*). Inspired by this, several recent research efforts have sought to incorporate "human gaze" to predict human drivers' attention (*20–22*) when performing driving tasks.

With regard to DLCV models in autonomous driving, several works have applied visual attention to explainable models (*23*)(*9*)(*24*). These systems generally represent an end-to-end process that can output the driving actions with attention maps displaying the "focus" of the model when generating such decisions. This can be observed in **Figure 1**, where a vehicle focuses its attention on the obstacles to its left, predicting that a left-turn is not feasible. Although these models have demonstrated superior performance in predicting driving actions together with interpretable attention maps, the question of adequate comprehension (by the model) of the details in the driving scenario remains unanswered. Specifically, regarding the prediction depicted in Fig.1, a clear problem is that there are two reasons why the vehicle cannot turn left: (a) there exist multiple vehicles to its left, and (b) there exists a double solid pavement marking line that prohibits left turn. While the model pays attention to the correct region to generate acceptable driving decisions, the exact rationale behind the decision remains ambiguous. Ideally, the model is expected to simultaneously learn both reasons. To achieve a holistic understanding of the driving scenario, the model needs to have the capability to explore exhaustively all the reasons for a driving action. This can be guaranteed by casting the autonomous driving task as an image-captioning and outputting the textual description (verbal rationale) of the observed driving situation (i.e., "obstacles in the left lane" + "solid line on the left" for **Figure 1** instead of directly outputting the driving decision). To this end, a joint modeling of both image and natural language is needed.

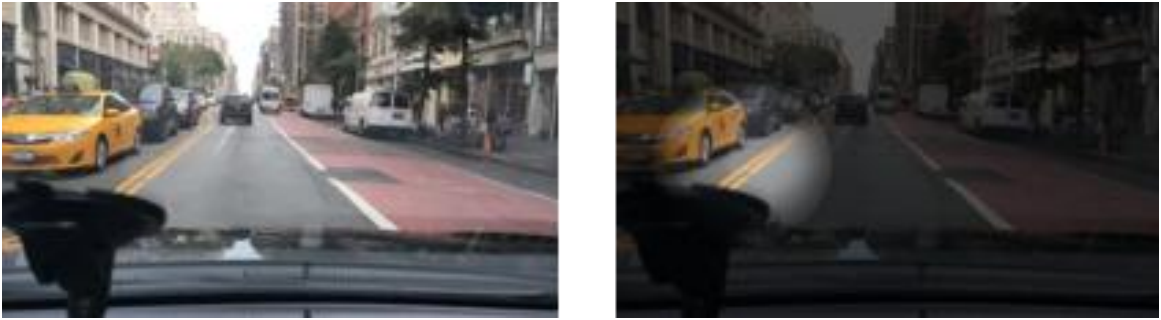

**Figure 1.** Raw image (left) and attended image (right). Bright region is the attention region (model focus). Predicted driving decision: cannot turn left

### B. Joint Modeling of Language and Image

From another perspective, during the training of a human driver, the trainer generally uses language to first guide the trainee's attention to look towards important objects and understand the scenario, and then to generate the driving decisions though simple causal relationships. For instance, the instruction "there is a solid line in your left, so change to the lane on the left is prohibited" will first lead the trainee to view the pavement lane markings on the lane to the left, and understand the driving scenario. Then, the corresponding driving decision "cannot move leftwards to the next lane" is the result based on the understanding of scenario, which demonstrates a clear causal relationship. In the first part this procedure (understanding the scenario), the trainee must simultaneously process both visual and verbal information and use the verbal instruction to guide the attention. After the training, the trainee is able to "self-attend" to the important regions without the trainer's instruction. A similar training process can be adopted using DL by jointly modeling the natural language and image while using language to induce the visual attention. Although this idea has been applied widely in image captioning models (*25*) and visual question answering (VQA) systems (*26*), it has not been fully explored in autonomous driving. Inspired by the success of this concept in other domains, we are motivated to translate the standard end-to-end autonomous driving task (maps image to driving decisions) into an image-captioning task (maps image to reasons) while applying similar language induced visual attention mechanism. More specifically, during the training, using the reason language to induce the attention on the image. During the testing, the model can perform attention automatically from the visual context and previously generated sentence.

After fully understanding the driving scenario by generating such reasons, the corresponding driving decisions can be made using causality rules. Compared to the end-to-end models, obtaining driving decisions through reasons prediction enables the system to reap the following benefits: 1) the potential to explore all the possible reasons for generating a driving action; 2) superior attention efficiency by associating attention regions to each word in the generated verbal reason. 3) superior interpretability for model diagnosis and improvements.

*C. Main Contribution*

In summary, this paper's main contributions are:
- Development of an end-to-end DLCV model that generates a verbal description and driving actions for AV system
- Successful imitation of human learning using a reason-induced attention mechanism
- Demonstrated the capability to ensure interpretability and remove ambiguity in driving scene comprehension through textual description and attention maps
- Demonstrated the capacity of the proposed model to expose current shortcomings and identify directions for improvements.

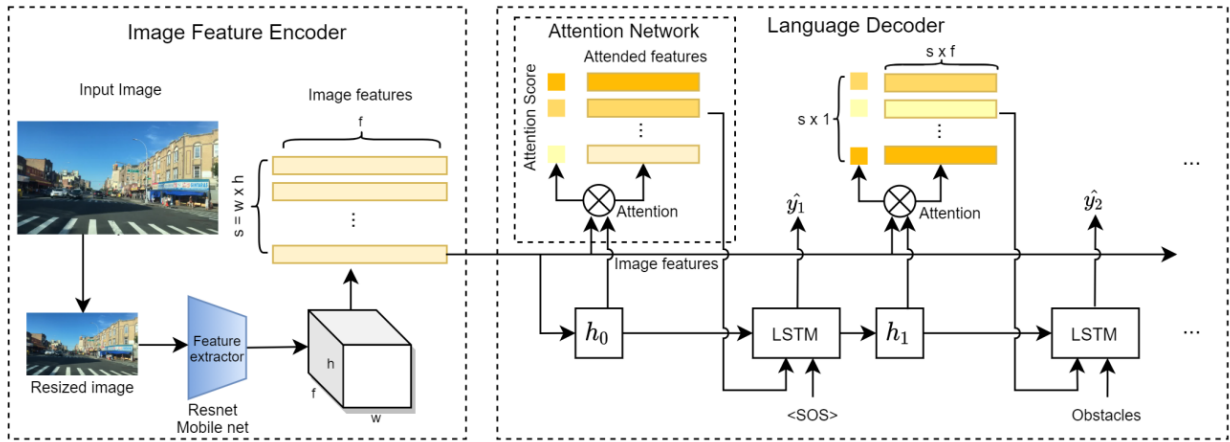

**Figure 2. Proposed model architecture**

**METHODS**

The proposed model architecture (Figure 2) adopts an encoder-decoder structure and has 2 blocks named Image Feature Encoder and Language Decoder.

*A. Image Feature Encoder*

As its name suggests, the Image Feature Encoder block has the function of extracting visual features from the input images. This is achieved by cascading a preprocessing layer (resize and normalization) and a classic pre-trained Convolutional Neural Networks (CNN) backbone model. Computationally, this block takes a standard RGB image with shape ($H \times W \times 3$) as input and output a 3D tensor with shape ($h \times w \times f$), where $f$ is the feature dimension of each spatial location and $h, w$ represent the height and width of the output feature map. For easier computation, the image features are flattened in the spatial dimension. The final output from the encoder is the image features $X \in \mathbb{R}^{s \times f}$ where $s = h \times w$.

In this paper, two backbone models namely Resnet50 (*27*) and Mobilenet_v2 (*28*) are tested. The reason for selecting these two backbones is the opportunity to consider the trade-off between accuracy and computation cost: while Resnet model is believed to be the state of the art in generating image

*Chen, Dong, Du, Li, and Labi*

features, Mobilnet_v2 is designed to run on portable devices with smaller number of parameters and much higher computational speed.

## B. Language Decoder

The function of the Language Decoder is to process the encoded image features and generate the descriptions of the scenario (reasons) word by word. In this regard, this module contains two structures: the attention network and language generator network. The former computes a vector of attention weights and "attended" image features. The latter structure uses classic Long Short-term Memory (LSTM) network to generate language sequences word by word.

To capture the correspondence between the language features and image features, and to imitate the human learning process, a soft attention mechanism as introduced in (*25*) is applied. Unlike the hard attention mechanism which only "selects" limited number of regions that are considered more important (giving zero weight to the unimportant regions), the soft attention mechanism conducts the feature selection through a weighted fusion process. More specifically, it first computes a probability distribution over the regions (attention weights) and uses this distribution to rescale the features. In this work, a "scaled dot product" attention is applied at each timestep $t$ (decoder generates one word for each timestep) with the following computations:

- Project the raw image features $X$ to the same embedding space as the hidden state of LSTM using a linear layer. The projected image feature is computed as:

$$x = XW \in \mathbb{R}^{s \times d} \tag{1}$$

where $d$ is the LSTM hidden state dimension and $W$ is the corresponding weight for the attention network.

- Conduct scaled dot-product between the hidden state of the previous time step $h_{t-1} \in \mathbb{R}^d$ and the projected image features $x$. Specifically, the output attention map $A_t$ for timestep $t$, is computed as:

$$A_t = softmax\left(\frac{x \cdot h_{t-1}}{\sqrt{d}}\right) \in \mathbb{R}^s \tag{2}$$

where $\sqrt{d}$ serves as a normalization constant to ensure numerical stability. Intuitively, this dot product computes a cosine similarity between the language feature and image feature. Therefore, the image region with higher attention weight is generally "closer to" the context in the language.

- Then the image is "attended" by computing the element-wise product between the attention map and raw image feature as follows:

$$x_t = A_t \circ x \in \mathbb{R}^{s \times d} \tag{3}$$

After attention, the rescaled image feature map $x_t$ is fused with the word embeddings of previous generated word $e_t$ through concatenation. This fused feature (containing both attended image information and language information) is then fed into LSTM cell to compute the hidden state $h_t \in \mathbb{R}^d$ at timestep $t$.

Finally, the language generation follows the standard logic in all seq2seq models: At each timestep $t$, pass the hidden state $h_t$ into a fully connected network (FCN) with the number of neurons in the output layer equals to the vocabulary size (vocabulary represents all the possible words the model can generate) to compute a vector of scores for all the words. Then, a $softmax$ function is applied to transfer the score vector into a probability distribution over words $p(\hat{y}_t)$.

*C. Loss Function*

Subsequent to the standard language modeling procedure, the model is trained end-to-end with a temporal multiclass cross-entropy loss, as follows:

$$L(y, \hat{y}) = -\frac{1}{N}\sum_{i=1}^{N}\sum_{t=1}^{T}\sum_{c=1}^{M} y_t \log(p(\hat{y}_t)) \qquad (4)$$

Where $M$ is the number of total classes (vocabulary size), $T$ is the total length of each sentence, and $N$ is the total number of training examples (training set size), $y_t$ and $\hat{y}_t$ are true label word and predicted word, respectively.

**EXPERIMENT SETTINGS**

*A. Dataset Preparation*

The dataset used to train and evaluate the model is a subset selected from the BDD Object Induced Actions (BDD-OIA) dataset proposed by (*9*). It extends the famous image based driving dataset BDD-100K (*29*) by labeling each frame individually with driving actions and explanations. The actions are the high-level feasible actions that can be undertaken by the driver at that specific time step. The explanations are associated with the actions and are summarized into 21 classes as shown in Table I. Fig. 3 presents examples of images and potential decisions. Since the dataset is unbalanced with some of the classes having images of size exceeding 10k while others are less than 20k, we select a subset containing six of the most frequent reasons: "**obstacles on the left lane**", "**no lane on the left**", "**solid line on the left**", "**obstacles on the right lane**", "**no lane on the right**", "**solid line on the right**". These six classes were picked because they are associated with left/right turn or lane-change actions, and these have been found to be associated with crashes (*30–33*). In addition, these actions require the model to learn a concept of direction while attending to "peripheral" regions of the images. For the selected six reasons, the corresponding total number of images are: 8475, 7989, 7625, 11521, 6771, 4750 respectively. Although this exhibits some imbalance, the number of images for each reason is large enough to train the model. In summary, the total number of images is 24,921, we further partition the dataset into training set (19936 images) and testing set (4985 images).

**TABLE 1 Actions With Explanations in BDD-OIA**

| Actions | Explanations |
|---|---|
| Move forward | Traffic light is green |
| | Follow traffic |
| | Road is clear |
| Stop/Slow down | Traffic light is red |
| | Traffic sign |
| | Obstacle: car |
| | Obstacle: person |
| | Obstacle: rider |
| | Obstacle: others |
| Turn left | No lane on the left |
| | Obstacles on the left lane |
| | Solid line on the left |
| | On the left-turn lane |
| | Traffic light allows |
| | Front car turning left |
| Turn right | No lane on the right |
| | Obstacles on the right lane |
| | Solid line on the right |
| | On the right-turn lane |
| | Traffic light allows |
| | Front car turning right |

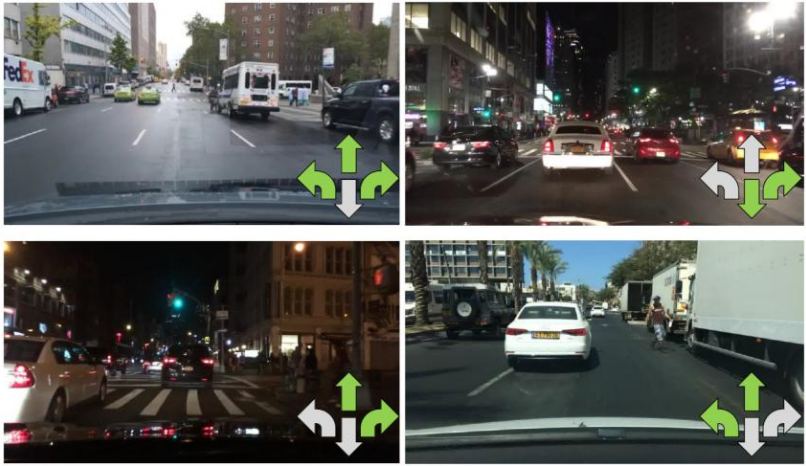

**Figure 3. Examples of images and decisions ground truth of BDD-OIA dataset (figure borrowed from (*9*))**

*B. Implementation Details*
*Language (reasons) preprocessing*
        In this study, we apply two NLP preprocessing techniques – tokenization and word embedding. Tokenization splits the sentence into words and assigns a unique token to each word. In addition to all the unique words (each word has a unique token), we added four extra tokens: <SOS> start of sentence; <EOS> end of sentence; <;> the delimiter of the reasons; <NULL> the placeholders after the <EOS> token. For each sentence, we first added the <SOS> and <EOS>, and delimited the reasons inside a sentence with <;>. For this work, different images can have different number of reasons. Further, the number of words for each reason is generally different. Therefore, this problem can be considered a dynamic length input-output problem with variable sequence length. To accommodate such discrepancy in the sequence length for all the images and train the model by batch, a padding manipulation is applied by adding <NULL> token to the positions after the <EOS> token. In this manner, the model can always accept and predict a fixed length sentence. Here, we use the maximum length of the possible sentence as this fixed length, which is the word count of total of six reasons plus the delimiter and start/end token.

        After tokenization, we applied a word embedding layer to transform the token representation to numerical vectors (word embeddings $e_t \in \mathbb{R}^d$). Initially, such numerical representation is generated from a standard Gaussian distribution; after training, it was observed that words with similar meaning tend to have "closer" representations on the embedding space.

*Other model details*
- Initial hidden state $h_0$ and initial context $c_0$: In the LSTM model, $h_0$ and $c_0$ represent knowledge prior to the language generation. In this work, $h_0$ and $c_0$ should contain the image information before any attention manipulation. Thus, they are computed with 2 separate FCNs using the mean of the image features: $c_0 = FC_1(mean(X))$; $h_0 = FC_2(mean(X))$.
- Input gate: As suggested in previous research (*25*), before passing the attended image features into the LSTM cell, one can add an input gate to control the quantity of visual information entering into the language decoder. The intuition is that there exist "helper words" such as conjunctions in natural language (for example, "is" and "the") whose function is to fulfill the requirement of grammar. These words generally do not describe the image, thus can be generated only based on the language context or previous generated sentence. With such input gate, the model has the ability to tune the dependency of the image features and language context. In the present study, the gate consists of a linear layer and sigmoid activation function, and takes the previous hidden state $h_{t-1}$ as input.

$$g_t(h_{t-1}) = sigmoid(Wh_{t-1}) \in [0,1]^d \qquad (5)$$

# RESULTS
## A. Evaluation Metric

The model seeks to comprehend the driving scene with the human language. Therefore, an ideal metric of performance should simultaneously consider the quality of generated language and correctness of the generated reasons. To evaluate the quality of generated language, we incorporate the classic evaluation metric in image captioning and neural translate, the Bilingual Evaluation Understudy (BLEU) score (*34*). This score measures the similarity between two sentences by computing the overlap between a hypothesis sentence and a reference sentence. Therefore, a higher BLEU score indicates better language generation quality since the generated sentence is closer to the ground truth label. In this work, we report the average BLEU scores for all images in the test set.

However, as mentioned in past research (*35*), the BLEU score itself is not able to capture the semantic meaning of the sentence, which may lead to poor correlation with human judgement. In this study, we compute the F1 scores for these two predictions to further evaluate the correctness of the generated reasons and actions. Specifically, the actions are predicted based on the reasons, and in this setting, there are two (infeasible) actions: "cannot turn left" and "cannot turn right" associated with the aforementioned six reasons. The action prediction follows the simple rule: if there exist any of the first 3 reasons, then the vehicle cannot turn left; and if there exist any of the last 3 reasons, then the vehicle cannot turn right.

More specifically, two versions of F1 score, namely overall F1 score ($F1_{all}$) and mean F1 score ($mF1$), for both reasons and actions are computed. $F1_{all}$ is the F1 score over all the predictions and is calculated as follows:

$$F1_{all} = \frac{1}{|A|} \sum_{j=1}^{|A|} F1(\widehat{A_j}, A_j) \qquad (6)$$

Where $\widehat{A_j}$ is the predicted value and $A_j$ is the true label (can be both reasons and actions), $|A|$ is the total number of predictions. Meanwhile, the dataset is still unbalanced, for example, there are more images for the "containing obstacles" reason compared to the "containing solid line" reason. Therefore, we calculated the F1 score within each predicted class, and each class, computed the in-class mean F1, as follows:

$$mF1 = \frac{1}{C} \sum_{j=1}^{C} \sum_{i=1}^{n} F1(\widehat{A_i^j}, A_i^j) \qquad (7)$$

Where C represents the number of classes in prediction (6 for the reasons and 2 for the actions).

## B. Comparative Analysis

To the best of authors' knowledge, this is the first work that casts the roadway scene interpretation in autonomous driving as an image-captioning task. As such, there currently exist no benchmark models for comparison with our proposed model. Therefore, we repeated the model development using different architectures and conducted comparative analysis across the models. We trained 3 models in total: the first 2 models are with different feature encoder (Mobilenet_v2 and Resnet50) to test the efficacy of backbone models; the third one is to test whether the input gate is helpful by incorporating the gate structure in an additional Resnet model.

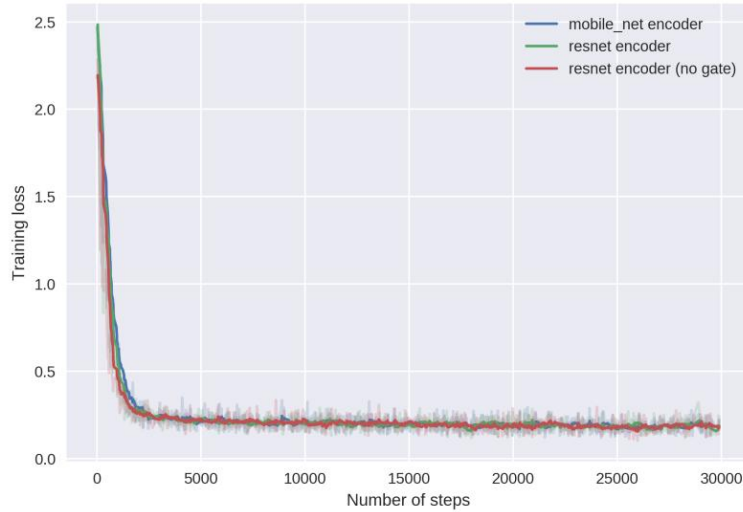

**Figure 4. Training curves for 3 models**

**Figure 4** presents the training curves for all the 3 models and **Table 2** presents their numerical evaluation metrics (Avg.BLEU score and F1 scores) for both reasons and actions. It was observed that the Resnet encoder model with no input gate has the highest performance in terms of model convergence rate during training and all the numerical evaluation metrics. However, there are little differences in performance across the models. This result suggests that at the phase of practical deployment of the model, the use of "lighter" encoders such as Mobilenet can achieve satisfactory performance with much higher computational efficiency. Also, it is observed that the input gate structure is not helpful for generating the reasons in this study. The potential reason lies in the labels of the dataset: each ground truth reason contains few "helper words" ("the", "on"), while a majority of the words are meaningful nouns and prepositions. In other words, generating majority words for this dataset requires the model to actually "look into" the image and obtain adequate visual information. Therefore, the gate structure cannot be sufficiently trained to improve the model performance. However, the effectiveness of this gate structure is expected in generating general sentences with more "helper words".

Another observation is that the model has high performance in the action prediction in terms of F1 scores. This can be attributed to generation of the actions from reasons, and multiple reasons can lead to the same action. Therefore, compared to the end-to-end image-to-action prediction models, the proposed model can add some redundancy in the actions prediction by first predicting the reasons; this further enhances the overall reliability of the driving system.

**TABLE 2. F-1 Scores of Proposed Model and Baselines**

| Models | Reasons Avg. BLEU score | Reasons $F1_{all}$ | Reasons inner-class $mF1$ | Actions $F1_{all}$ | Actions inner-class $mF1$ |
|---|---|---|---|---|---|
| Mobilent | 0.478 | 0.521 | 0.682 | 0.917 | 0.847 |
| Resnet | 0.473 | 0.529 | 0.678 | 0.917 | 0.847 |
| Resnet (no gate) | **0.497** | **0.538** | **0.702** | **0.917** | **0.848** |

*Chen, Dong, Du, Li, and Labi*

## C. Visualization of Attention

**Figure 5** presents two sample predictions from test set, which contain the raw image and the attention map for each word. As expected, the model was found to be capable of generating legal sentences with correct words and grammar to describe the driving scenes.

These results demonstrate the model can indeed capture the semantic relationship between the image regions and language. For example, in the first image, there exist vehicles on the both sides of the ego-vehicle. Therefore, the model's attention is guided specifically to those regions, and then is capable of predicting the correct reasons and actions. In the second image of **Figure 5**, the model has successfully learned both reasons ("obstacles" and "solid line") as expected in the left region. One observation from the attention map is that the model tends to look into both directions when predicting directional words "left" and "right". We believe this could be attributed to the fact that these directional concepts are relative, therefore, the model can understand the word "left" by simultaneously understanding which direction is "right".

From the perspective of explainable models, the attention maps can help the developers understanding the model behaviors and shortcomings. For example, in this work, a drawback for the model (demonstrated in both figures in **Figure 5**) is that it does not possess a "concept" of distance, i.e., the model can attend to a majority of the obstacles in the vicinity, but cannot understand that only the obstacles in very close proximity should be attended for generating the actions. This can be mitigated using a finer-labelled dataset which has the labels for objects that are "closer" and those that are "further" and then, training the model to distinguish between objects and obstacles based on their proximity.

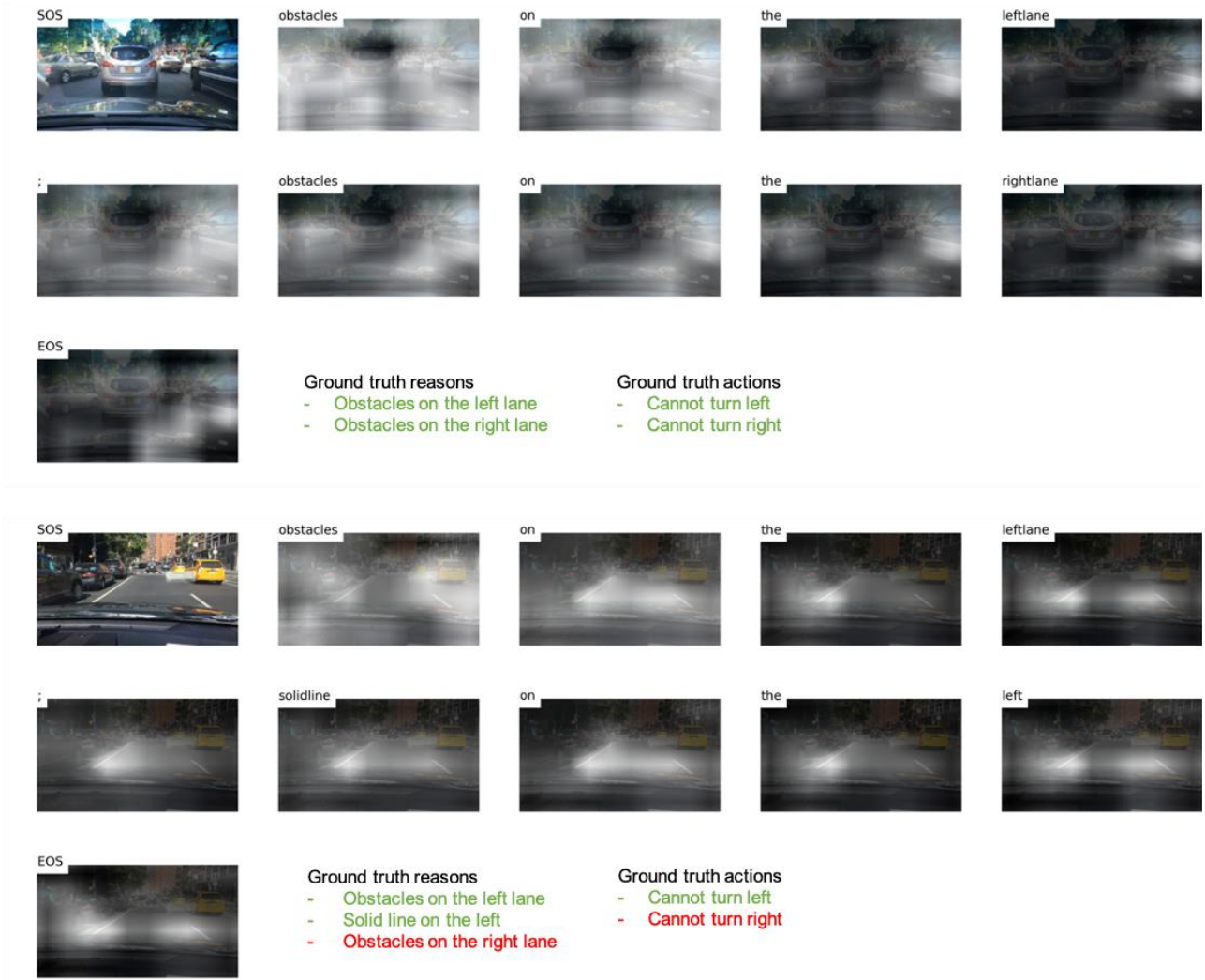

**Figure 5.   Example predictions. Brighter part indicates the attention region for each word in the generated reason. For both reasons green color indicates true positive, red is the false negative**

In summary, by visualizing the attention maps and generated reasons, the model behavior can be understood. Further, this process can be used to identify the weaknesses of the currently trained model and to seek the potential directions for improvements. These extra benefits in interpretability and diagnosability generally do not exist in majority of other DL models that exist in the literature.

## CONCLUSION

In this paper, an explainable DL model with reason-induced visual attention is proposed to mitigate the low interpretability problem associated with deep neural networks while ensuring a thorough understanding of the driving scene. The results from training and testing process suggest that the proposed model the potential to imitate the attention mechanism of human drivers by jointly modeling the image and language inputs while using the language to induce attention. Further, intermediate outcome and attention maps are shown to be helpful in facilitating an understanding of the model behavior and in identifying the model drawbacks and potential solutions.

Moving forward, the model trained in this work can be improved upon by training on better-labeled datasets that include directions and distances. In addition, with the development of the sibling technology: vehicle-to-everything (V2X) connectivity (*36–42*), reason-induced visual attention can be extended to attention between other information sources. For example, using the image perception knowledge (i.e., vehicle heading) to attend the connectivity information (i.e., speed, location, acceleration of surrounding vehicles) to extract the relevant information (i.e., only investigate the vehicles in that direction).

Ultimately, explainable autonomous driving models are expected to have significant applications during deployment of AVs. Knowing the rationale behind an AV's actions will be consequential in situations that require thorough liability analysis. For example, in the event of a collision between an AV and a human-driven vehicle, the liable parties will be easily identified, and the causality of collision can be clearly attributed (human error, improper detection, etc.). Increasing the explainability of autonomous driving decisions can help contribute to enhanced reliability of AVs, which, in turn, may improve public perception and trust of AVs.

## ACKNOWLEDGMENTS

This work was supported by Purdue University's Center for Connected and Automated Transportation (CCAT), a part of the larger CCAT consortium, a USDOT Region 5 University Transportation Center funded by the U.S. Department of Transportation, Award #69A3551747105. The contents of this paper reflect the views of the authors, who are responsible for the facts and the accuracy of the data presented herein, and do not necessarily reflect the official views or policies of the sponsoring organization.
This manuscript is herein submitted for PRESENTATION ONLY at the 2022 Annual Meeting of the Transportation Research Board. A full version of the paper is currently under review by another journal.

## AUTHOR CONTRIBUTIONS
The authors confirm contribution to the paper as follows: all authors contributed to all sections. All authors reviewed the results and approved the final version of the manuscript.

*Chen, Dong, Du, Li, and Labi*